\documentclass[a4paper]{article}
\usepackage[utf8]{inputenc}
\usepackage[english]{babel}  % Standard LaTeX language package
\usepackage{amsmath,amssymb}
\usepackage[numbers]{natbib}
\usepackage{emnlp2018}
\usepackage{subfig}
\usepackage{multirow}
\usepackage{graphicx,color}
\aclfinalcopy
\usepackage[colorinlistoftodos]{todonotes}

\title{\textbf{Language Independent Named Entity Recognition via Orthogonal Transformation of Word Vectors}}
\author{\textbf{Omar E. Rakha}\hspace{1cm} \textbf{Hazem M. Abbas}\\
\textbf{Dept. Computer and Systems Engineering}\\
\textbf{Faculty of Engineering}\\
\textbf{Ain Shams University, Cairo 11571, Egypt}
}
\date{}

\begin{document}
\maketitle
\begin{abstract}
%% Text of abstract
Word embeddings have been a key building block for NLP in which models relied heavily on word embeddings in many different tasks. In this paper, a model is proposed based on using Bidirectional LSTM/CRF with word embeddings to perform named entity recognition for any language. This is done  by training a model on a source language (English) and transforming word embeddings from the target language into word embeddings of the source language by using an orthogonal linear transformation matrix. Evaluation of the model shows that by training a model on an English dataset the model was capable of detecting named entities in an Arabic dataset without neither training or fine tuning the model on an Arabic language dataset.
\end{abstract}

% \linenumbers
\section{Introduction}

Named Entity Recognition (NER) is an important component of information extraction that aims at extracting named entities from textual data. Named entities are words that represent a known person, location, organization or any other entity that can be identified by its name. Extraction of named entities can be used to further improve search engine queries \cite{guo2009named} or for question answering \cite{molla2007named}. Detection of named entities requires building hand annotated datasets for each target language which consumes significant human labor and time.

NER systems used to rely on hand coded features, part-of-speech (POS) tags, semantic lexicons and huge gazetteers \cite{benajiba2007anersys}. However this restricted the model's performance and required huge work and domain knowledge to find suitable features.

Current state-of-the-art systems in NER are very accurate with performance exceeding 90\% F-score \cite{huang2015bidirectional} \cite{lample2016neural} by using a neural network composed of a Bidirectional LSTM layer \cite{graves2013speech} and a Conditional Random Field (CRF) classifier \cite{lafferty2001conditional}. However, these models are only trained for a specific language (English for example) and can not be used on any other language.

Due to language dependence, extending the knowledge obtained by these models to new languages is impossible. Thus trying to support new languages requires building new models that only work on the language they are trained on. The effect of this is that the number of languages that can be supported in an application is limited by the amount of labeled data that can be obtained in every language and the performance of each model.

In this work, a novel method is proposed for NER by training a model on an English corpus and evaluating the resulting model on an Arabic corpus by transforming the Word embedding space from Arabic to English. The method can be extended to any language as long as it is possible to create the transformation matrix for its word embeddings. The proposed model does not  explicitly rely on language specific features since the employed word embedding model implicitly captures language specific features and benefits from morphological features of words. This brings about a model that is both simple and efficient.

The paper organization is as follows. The problem is defined in details in Section \ref{sec2}. Previous NER models are reviewed in Section \ref{sec3}. The proposed language independent NER model is presented in Section \ref{sec4}. Experimental results produced when the model is applied to a new target language are analyzed in Section \ref{sec5} and Section \ref{sec6} concludes the paper. 

\section {Problem Statement} \label {sec2}
The language independent NER problem can be composed of two sub problems the first being the detection of named entities and the second being language independence.
The sub problem of detecting named entities is a sequence labeling problem in which the goal is to find out words that are named entities (e.g.,: Persons, Organizations, Locations, ...).

A named entity is a word or a phrase that stands consistently for some referent. This includes names of people, places, organizations. This can also include temporal expressions (e.g., January fifth 2010, August, 20 November). Numerical expressions can also be considered named entities (e.g., \$50, 25.5\%).

\begin{figure}[h]
    \centering
    \includegraphics[scale=0.3]{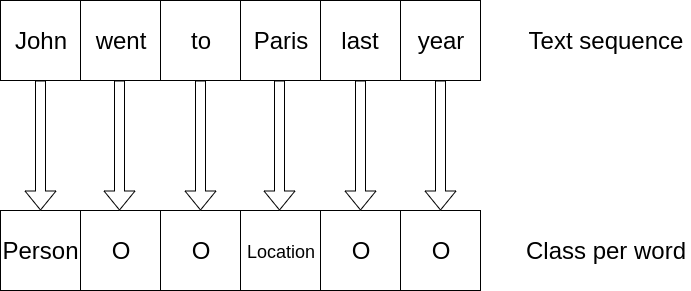}
    \caption{An NER Example}
    \label{fig:ner-example}
\end{figure}

\begin{figure}[h]
    \centering
    \includegraphics[scale=0.25]{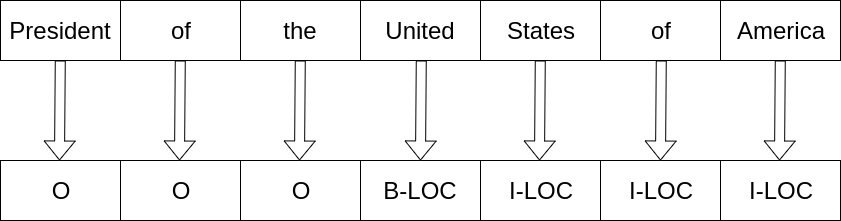}
    \caption{The IOB format}
    \label{fig:ner-iob-example}
\end{figure}

For example in Figure \ref{fig:ner-example} each word is labeled in a text sequence as a named entity (In this case a person and a location).
Outside words are labeled with "O" following the IOB (Inside-Outside-Beginning) format presented in \cite{DBLP:journals/corr/cmp-lg-9505040}. The IOB format prefixes the class of the beginning word with a "B-" indicating it's the first word in this named entity. while the following words that belong to the same entity will be prefixed with "I-" which stands for inside. This way we can label named entities that are longer than 1 word. Refer to Figure \ref{fig:ner-iob-example} for illustration.

The second sub problem addressed here is language independence. The goal of building language independent models is to have solutions that can ultimately solve problems regardless of the language it receives. Ideally these models will learn the underlying structure of human language and thus be able to tackle the problems addressed at it without being concerned with the language it uses.

\begin{figure}[h]
    \centering
    \includegraphics[scale=0.30]{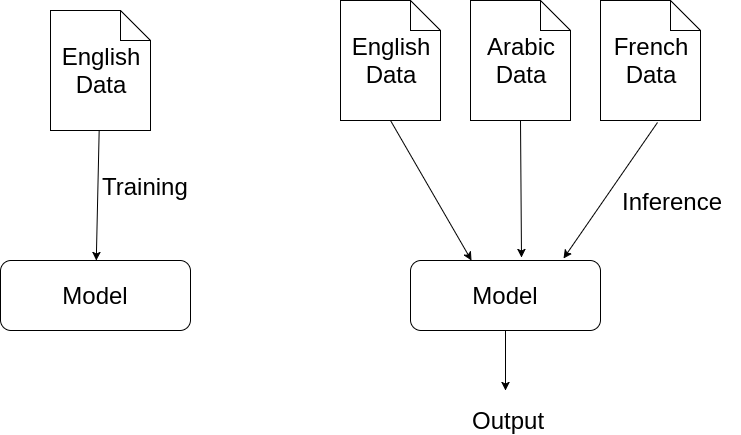}
    \caption{A language independent model}
    \label{fig:ner-lang-independent}
\end{figure}

For example there exist many data sets for NER in English. However, Arabic data is very scarce. A model that is trained using English data and can be extended to solve Arabic input is said to be language independent (Figure \ref{fig:ner-lang-independent}).
\section{Related Work} \label{sec3}
Currently NER multilingual systems approach the problem by relying on data sets that are designed for building multilingual models like the UN parallel corpora \cite{ziemski2016united}. In following subsections we will explore these approaches and their limitations before proposing the new approach.
\subsection{Cross Lingual Resources}
The cross lingual resources approach is centered around finding common features in languages or features that are language dependent. The goal of this approach is to unify the features of more than one language in order to build a classifier that is capable of finding the named entities based only on these features.

The system \cite{darwish2013named} incorporates Wikipedia and other knowledge bases like DBpedia \cite{jl_2014/swj_dbpedia}. It uses features like cross-lingual capitalization, transliteration and DBpedia based labeling.
% \begin{itemize}
% \item Cross-lingual Capitalization
% \item Transliteration
% \item DBpedia based labeling
% \end{itemize}
\\

\textbf{Cross-lingual Capitalization}: Since Arabic has no capitalization DBpedia was used to find capitalization weights for a word and be used as a feature.

An example: the phrase ”Pacific Ocean" was capitalized 36.7\% of the time. Thus, the arabic word for "Ocean" was assigned the feature "B-CAPS-0.4" and the arabic word for "Pacific" was assigned the feature "I-CAPS-0.4". The prefix of the feature determines if the word was a "Beginning" or "Inside" word. The body of the feature "CAPS" symbols capitalization and the suffix of the feature is the frequency of being "CAPS".

\textbf{Transliteration}: The intuition behind transliteration is that named entities are usually transliterated rather than translated. For example the word "Hassan" is an arabic name and also means "good". They use a transliteration weight as a feature similar to the capitalization weight.

\textbf{DBpedia based labeling}: DBpedia is a large collaboratively-built knowledge base in which structured information is extracted from Wikipedia. For example, NASA is assigned the following types: Agent, Organization, and Government Agency. These types were used as a feature for each word (if it exists in DBpedia).

The problem of this approach is the heavy reliance on the nature of a language, i.e., it has capitalization, transliteration is valid, and it is well represented in DBpedia.

\subsection{Multi Task Cross Lingual Training}
Due to the limitations of the previous approach in having to find features that are viable in more than one language, like Capitalization, a better approach was needed to overcome this limitation. A suitable approach was to rely on embedding to let the model find these features by itself.

In multi task cross lingual training \cite{yang2016multi}, a model is trained on multiple tasks such as POS, chunking and NER for more than one language at the same time by using a character level embedding GRU \cite{chung2014empirical} model and a word level embedding model. This model is trained with a shared embedding layer to capture the morphological similarity between languages and learn word vectors that are based on the two languages. This embedding space will capture the similarities between words from the two languages. The final layer is a Conditional Random Field (CRF) classifier \cite{lafferty2001conditional} which has specific weights for each task.

\begin{figure}[h]
    \centering
    \includegraphics[scale=0.25]{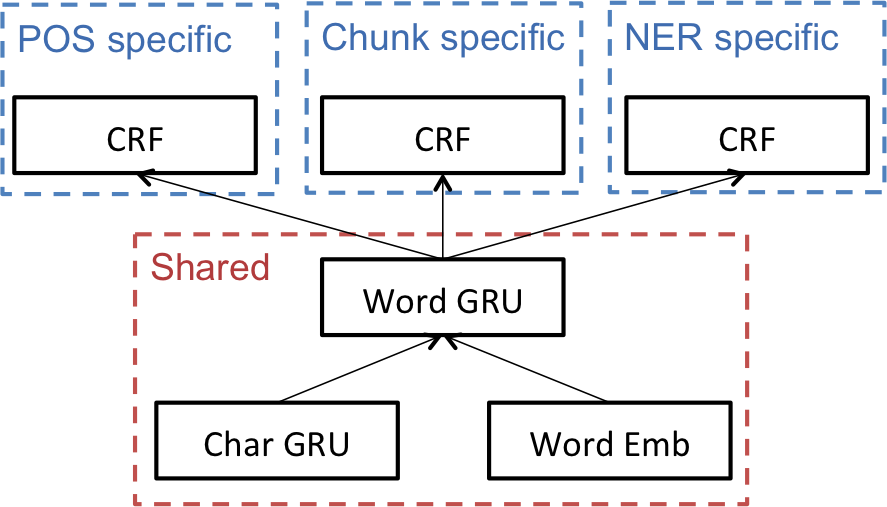}
    \caption{The Multi Task Cross Lingual Training model}
    \label{fig:multi-task}
\end{figure}

The model in (Figure \ref{fig:multi-task}) is structured as pluggable components. The bottom component is an embedding layer. This layer is shared among different tasks and different languages, which means that the model will use the same embedding layer if it was trained for English NER, English POS, Arabic Chunking or Spanish POS.
The preceding layer is a task dependent CRF \cite{lafferty2001conditional} classifier that works on top of the shared embedding layer to produce task specific results (for example: named entities in case of NER).

However this model will require a huge amount of annotated data in every needed language. In addition, the model will require retraining whenever more languages need to be added. Also the parameters of the model will increase significantly with the number of languages and hence increase the complexity of the model. The shared embedding layer will not be able to capture any similarity between languages that are intrinsically different like English and Arabic.

These approaches show promising results on the multilingual NLP tasks, however the growing complexity of these models with the number of languages makes it near impossible to have models that can work on a huge number of languages. Building a model to work with a few languages would consume a considerable amount of resources(time, annotation effort, ...). Therefore an approach that would work independently of the number of languages is needed.
\section{Proposed Approach}
\label{sec4}
In this section, an approach to build a model that is both efficient and can scale reliably with the number of languages is presented.
The idea of this approach is to build a neural NER model using the conventional components for the problem: word embedding for embedding words into vectors, Bidirectional LSTM for feature extraction and a Conditional Random Field sequence classifier.

\begin{figure}[h]
    \centering
    \includegraphics[scale=0.3]{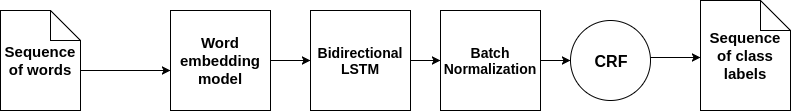}
    \caption{Architecture of the proposed model }
    \label{fig:model}
\end{figure}

In order to incorporate the language independence in the proposed model, an 
additional module, \textit{word embedding transformation}, is added to the pipeline. The main function of the transformation module is to transform the vectors of the words of a certain target language (Arabic in our case) into the space of the vectors of the source language (English in our case) for which the model was trained on. By doing so, the model will become independent of the language of the input and will only depend on the quality of the \textit{word embedding transformation} technique that is applied. 
Thus, the model pipeline will be composed of four modules:
\begin{enumerate}
\item Word Embedding
\item Word Embedding Transformation
\item Bidirectional LSTM Layer
\item Conditional Random Field
\end{enumerate}
Figure \ref{fig:model} shows the architecture of the proposed neural architecture. First, each word in a sentence is embedded  into its corresponding word vector using the word embedding model. If the word vectors are in a language different than the source language (English in this case) the word embedding transformation is invoked to transform the vector from the space of the target language to the space of the source language. The resulting word vector sequence is then used as the input to a bidirectional LSTM layer which encodes the input sequence. Finally a CRF layer is used to tag the sequence.

\subsection{Word Embedding}
Because language is a sequence of words, while neural networks only work with numbers we have to transform our words into a numerical format. Vector space models are the numerical representation of documents in which a vector is used as a representation for this document. An example of this is Bag of Words(BOW) \cite{harris1954distributional}.

In BOW models a vector is assigned for each document representing word counts for each word in the vocabulary in this document. Consider our vocabulary to only have the words: ["the", "boy", "girl", "rides", "bus", "and", "from"]. A representation of the string "the boy rides the bus and the girl rides the bus" would be [4, 1, 1, 2, 2, 1, 0] which is the count of each word in the vocabulary found in this string.

Vectors for words can also be computed using BOW by having a one hot representation over the vocabulary. For example the word "boy" would receive the vector [0, 1, 0, 0, 0, 0, 0]. this way we can have vectors representing every word in our vocabulary.

The limitation of this approach is that vectors are very sparse since for a word the vector will be mostly zeros except for a single value that corresponds to the word. These vectors also lack any information on the semantic meanings of these words since the position of the word in the vocabulary gives no information about similar words to it.

Due to these limitations distributed representation of words outperforms traditional BOW methods because they are dense vectors that can learn semantic and syntactic properties of these words.

\begin{figure}[h]
    \centering
    \includegraphics[scale=0.20]{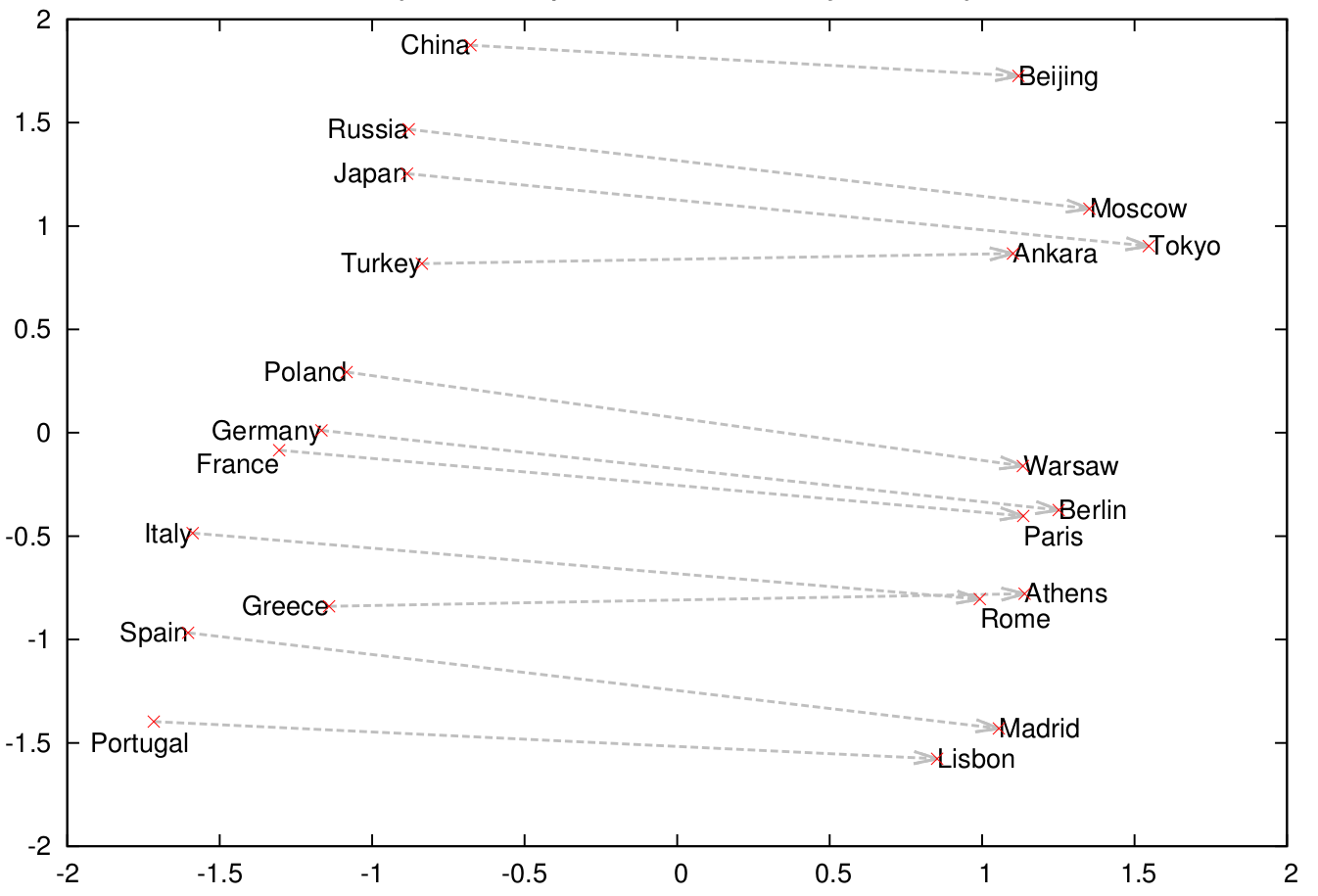}
    \caption{Country-Capital Relationship}
    \label{fig:country-Copy}
\end{figure}

A famous example for this taken from \cite{mikolov2013distributed} is how their model learned a relationship between Country-Capital in the PCA projection of these vectors (Figure \ref{fig:country-Copy}).

 From the 2 dimensional PCA projection we can see that the model learned similar representation for countries (China, Russia, Japan, ...) and similar representation for capitals (Beijing, Moscow, Tokyo, ...) and the direction of the vector between a country-capital is similar. For example the relationship $model("China") - model("Beijing") \approx model("Russia") - model("Moscow")$ can be seen from the projection.

These models are trained on huge corpora of data -Wikipedia for example- and the model learns from the context information of words the best representation that would capture semantic and syntactic information and embed it into the highly dense vector.

The Word embedding model used in the proposed approach is introduced in \cite{bojanowski2016enriching} which is a Skip Gram model \cite{mikolov2013distributed} with subword information. Given a matrix $\mathbf{W}$,  an entry $\mathbf{w}^i$ is the $i$th vector in the matrix which is the vector representation of the word $i$.
The Skip Gram's objective is to find word representations that are useful for predicting the surrounding words in a sentence or a document. The model is trained to maximize the average log probability over a sequence of words $\mathbf{w_1, w_2, w_3, ...., w_T}$ where $c$ is the size of the training context (training window) and $\mathbf{w}_t$ is the center word.:
\begin{equation}
\frac{1}{T}\sum_{t=1}^T\sum_{-c \leq j \leq c, j \neq 0} \log p(\mathbf{w}_{t+j} | \mathbf{w}_t)
\end{equation}
The basic skip gram model defines the probability of word $t+j$ given word $t$ as $p(\mathbf{w}_{t+j} | \mathbf{w}_t)$ using the softmax function:
\begin{equation}
log(p(\mathbf{w}_O | \mathbf{w}_I)) = log(\frac{exp(v^{'\top}_{\mathbf{w}_O}v_{\mathbf{w}_I})}{\sum_{w=1}^W exp(v_w^{' \top} v_{w_I})})
\end{equation}
where $v_w$ and $v_w^{'}$ are the input and output vector representations of $\mathbf{w}$ and $W$ is the number of words in the vocabulary.
This can be simplified into:
\begin{equation}
v^{'\top}_{\mathbf{w}_O}v_{\mathbf{w}_I} - log(\sum_{w=1}^W exp(v_w^{' \top} v_{w_I}))
\end{equation}
which consists of two terms, the dot product $v^{'\top}_{\mathbf{w}_O}v_{\mathbf{w}_I}$ which represents the similarity between the two vectors and the other term $log(\sum_{w=1}^W exp(v_w^{' \top} v_{w_I}))$ being the normalization term. Computing the normalization term is very expensive since it requires summing over all words in the vocabulary. This makes it impractical to compute the gradient of the softmax term $\bigtriangledown p(\mathbf{w}_O | \mathbf{w}_I)$ which is proportional to the size of the vocabulary $W$.

Negative sampling (NEG) is used as the objective for training the Skip Gram model instead of the softmax to overcome the problem of computing the normalization factor. NEG is defined as:
\begin{equation}
\log \thinspace \sigma(v_{w_O}^{' \top} v_{w_I}) + \sum_{i=1}^{k} \mathop{\mathbb{E}}_{\mathbf{w}_i} \sim P_{n}(w) [log \thinspace \sigma(-v^{' \top}_{w_i} v_{w_i})]
\end{equation}
Negative sampling works by randomly selecting a set of k "negative" words drawn from a uni-gram distribution $(P_{n})$, These words are the words the model should not be predicting and instead of updating the weights for every word in the vocabulary that is not the target word, the optimizer will only update weights for this negative sample avoiding a huge computation that would make model training in-feasible.

In other words the vectors are trained by teaching the model to differentiate between the target word and noise.

So far the model is only capable of learning a distinct vector representation per word. However, this ignores the morphological features of words (prefixes, suffixes, n-grams, ...). Therefore, the scoring function needs to be adjusted to take this information into account.

A proper modification for the model would be to learn vector representations for words and for character n-grams instead of learning vectors for words only. The implication of this is that each word will be decomposed into a set of n-grams and the word itself, each element of this set has a distinct vector. this allows for n-gram vector sharing between words thus producing high quality vectors that take into account the structure of the word itself.

For $\varrho_w \subset {1, ..., G}$ with G being the set of n-grams appearing in $\mathbf{W}$ and $\varrho_w$ being the set of n-grams appearing in word w, a vector representation is associated with each n-gram, $g$, a word is represented by the sum of the vectors of its n-grams and a unique vector for the word itself, the scoring function is therefore:
\begin{equation}
s(w,c) = \sum_{g \in \varrho_w} z_{g}^{\top} v_c
\end{equation}
Where $z_{g}$ is a vector representation of each n-gram g.

For example the word "Good" can be represented by the set of character n-grams of length 3,4 {"Goo", "ood", "Good"} where every n-gram of which will have its own vector representation.

\subsection{Word Embedding Transformation}
\begin{figure}[htp] 
    \centering
    \subfloat[An example of two English words in their embedding space]{%
        \includegraphics[width=0.3\textwidth]{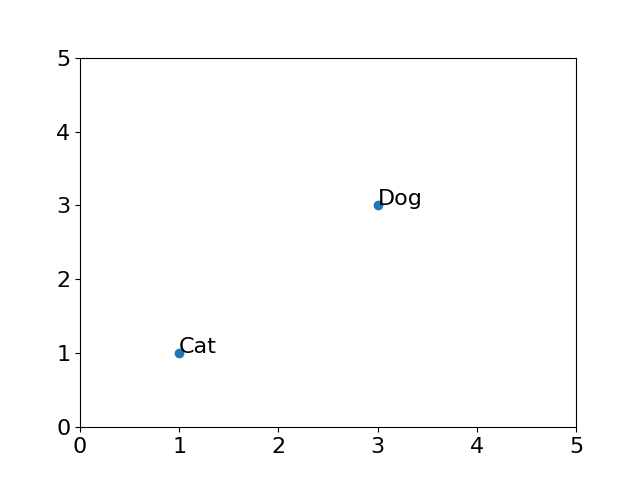}%
        \label{fig:a}%
        }%
    \hfill%
    \subfloat[An example of their Arabic translation in their embedding space]{%
        \includegraphics[width=0.3\textwidth]{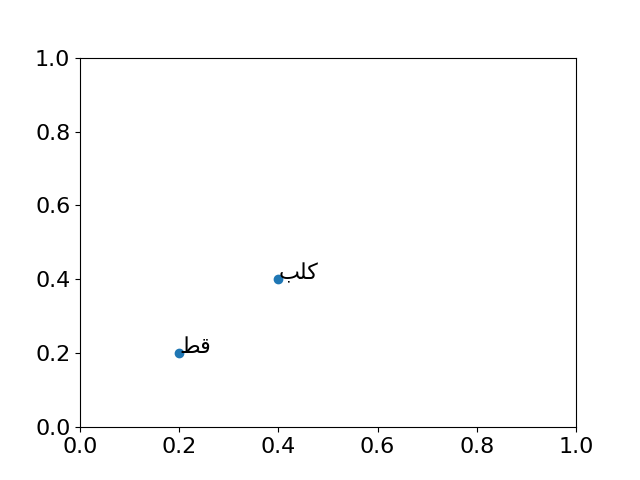}%
        \label{fig:b}%
        }%
    \hfill%
    \subfloat[Ideally after the transformation the two Arabic words will 
overlap 
with the two English words]{%
        \includegraphics[width=0.3\textwidth]{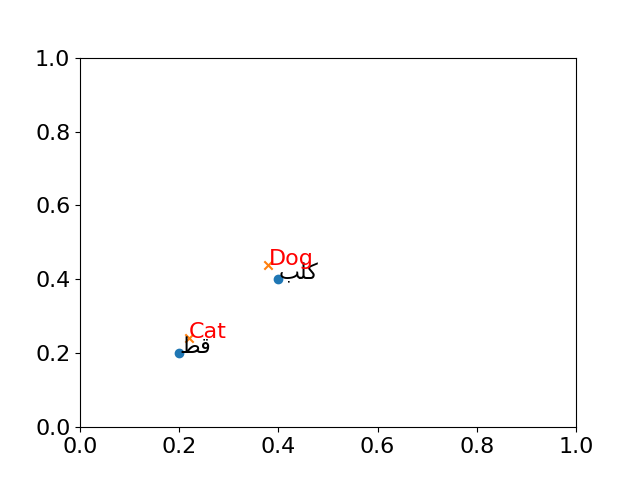}%
        \label{fig:c}%
        }%
    \caption{Word embedding transformation}
    \label{fig7}
\end{figure}

Figure \ref{fig7}  will help in explaining the idea of word embedding transformation.
In Fig. \ref{fig:a}, two English words are projected on a two dimensional space using a PCA projection of their word vectors. While in Figure \ref{fig:b} the same two words in Arabic are projected on their respective embedding space. It is quite clear that the two embedding spaces are entirely different and that the two words do not overlap(The position of the same word across different embeddings does not hold). The goal of word embedding transformation is to find a feasible transformation that can transform the vectors of one of the two spaces into the other space in order to have words with the same meaning ideally overlapping Figure \ref{fig:c}.

The basic objective of the proposed model is then to be trained on a corpus from a source language (English) (Fig. \ref{fig:a}), for which there are available annotated data, and test the model on a target language with less data (Arabic) (Fig. \ref{fig:b}).  
This would not be achieved by simply using the word embedding of the two languages separately. due to the fact that both languages have different distributions and the model will not be able to capture  the target word embedding. To circumvent this problem, we adopt the approach suggested in \cite{smith2017offline} to transform the word embedding of the target language to the source language and then use the latter in the proposed architecture as illustrated in Fig. \ref{fig:c}. This approach is tested by training the model on an English dataset \cite{nothman2012:artint:wikiner} and then evaluate the model on an Arabic dataset \cite{benajiba2007anersys}.

The transformation can be performed by using a translation matrix that is capable of approximating target word embedding into source word embedding. 
Translation matrix is a powerful approach that will map one matrix to another by matrix multiplication.\\
Suppose that there are  two sets, $X,Y$,  of word embeddings that correspond to the target and source languages, respectively, where $X,Y \in R^{n*d}$  with $n$ being the number of vectors and $d$ being the dimension of each vector.  The  goal is to find a matrix $W$ that will minimize the error function:\
\begin{equation}
\sum_{i=1}^{n} ||Wx_i - y_i|| ^ 2,
\end{equation}
where $x_i$ is our target language vector and $y_i$ is the source language 
vector for the $ith$ word.
One can employ the Stochastic Gradient Descent (SGD) optimization to reach the 
optimal 
matrix $W^{*}$ that minimizes the error function. Hence, any vector from our 
target language $X$ can be linearly transformed to a vector in the source 
language $Y$,   by 
multiplication with $W^{*}$.

The SGD  approach is normally susceptible to reaching sub optimal solutions by 
getting stuck in a  local 
minimum. Moreover, it will have to iterate over the  full dataset multiple times 
and thus consume computing power. \\
An alternative approach is to solve for the translation matrix discussed above directly which would produce an exact solution \cite{smith2017offline}.

First, let's assume that our word vectors in both spaces $\boldsymbol{Y_{n\times d}}$ and $\boldsymbol{X_{n\times d}}$ are normalized to unit norm in $l_2$. Therefore we can define the cosine similarity between two vectors as the inner product of two vectors $cos(y_i, x_i) = y_i^T.x_i$. Then, the similarity matrix between every pair of vectors in our two spaces will be $S = Y.X^T$ in which ever term corresponds to the cosine similarity between the two corresponding vectors.

In the default setting, we do not expect the similarity measures to be high. since the two spaces are not aligned yet. therefore we can modify our objective function to maximize the overall similarity of every pair of vectors representing the same word after alignment. let's assume we have a matrix $O$ which is our orthogonal transformation matrix, we will prove shortly why it has to be orthogonal.

\begin{equation}\label{eq:similarity-objective}
    \max_O \sum_{i=1}^n y_i^TOx_i
\end{equation}

As we find the value of $O$ that maximizes equation \ref{eq:similarity-objective} we are basically increasing the cosine similarity between two word vectors that represent the same word but in two different languages. Therefore our word vectors are aligned using the transformation matrix to be in the same space. In order to do this we first need to form two ordered matrices $X_D$ and $Y_D$ which are formed from a dictionary or an aligned corpus in which the $i^{th}$ row corresponds to the source and target language word vectors of the same word.
The first step is to multiply these two matrices in order to produce the similarity matrix $M = Y_DX_D^T$, then we take the SVD of $M$ which is $M = U \Sigma V^T$. Using the previously computed matrices $U$ and $V^T$ we can map both languages into a single space, by applying the transformation $V^T$ to the matrix $X$ and the transformation $U^T$ to the matrix $Y$. Therefore the similarity matrix for the aligned spaces will be $S = YUV^TX = y_i^TUV^Tx_j = (U^Ty_i)(V^Tx_j)$. Now that we have solved equation \ref{eq:similarity-objective}, let's prove the orthogonality constraint.

It is argued that the translation matrix \textit{must} be an orthogonal matrix 
\cite{xing2015normalized}. To show that let us first form the similarity matrix 
$S = Y(WX^T)$ where \textbf{Y} represents the word vectors matrix of our source 
language and \textbf{X} represents the word vectors matrix of our target language, the 
matrix \textbf{W} is the transformation matrix. 
Thus, the matrix \textbf{S} represents the dot product between every vector in 
the matrix \textbf{Y} and its corresponding vector in the transformed matrix $\boldsymbol{WX^T}$. For the matrices, $\boldsymbol{Y_{n\times d}.W_{d\times d}.X_{n\times d}}$, it can be noted that the similarity matrix's dimensions will be $S_{n\times n}$.  This matrix provides the similarity of each vector in the target language with each vector in the source language. Likewise, a reverse similarity matrix, $Q$ between the source vectors and the target vectors can be established as $S' = XQY^T$.
For this to remain self consistent, and because the two similarity matrices provide the similarity between the two 
matrices \textbf{X} and \textbf{Y}, it is evident that  $S' = S^T$. Since  $S^T = XW^TY^T$, 
therefore $Q = W^T$ which means that if $W$ maps the source language to the target 
language then $W^T$ maps the target language back to the source language.

Given the matrix $W$ we assume that $x \approx W^Ty$ and $y \approx Wx$ 
therefore $x \approx W^TWx$ therefore we can conclude that the transformation matrix $W$ must be an orthogonal matrix (or unitary matrix) satisfying $W^TW = I$, where $I$ denotes the  identity matrix.

\subsection{Bi-LSTM encoding}
Neural Networks have always had a strong ability to perform feature extraction even with time series or sequences they're still able to capture time related features due to their hierarchical nature and recurrence relations.

For this we will be using the Long-short term memory \cite{hochreiter1997long} algorithm to perform feature extraction from our word vectors sequence in order to convey temporal features to the classifier.

\begin{figure}[h]
    \centering
    \includegraphics[scale=0.4]{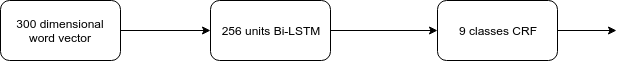}
    \caption{Model detailed architecture for each time step}
    \label{fig:network-architecture}
\end{figure}

LSTM is the core of our model (Figure \ref{fig:network-architecture}) since it will be responsible for extracting the useful features from the word vectors to help the classifier find the correct tags for the sequence.

The main strength of LSTM is its ability to handle long term dependencies in a sequence of steps while updating its state with information from context due to its gates that are responsible for updating cell states.

\begin{figure}[h]
    \centering
    \includegraphics[scale=0.23]{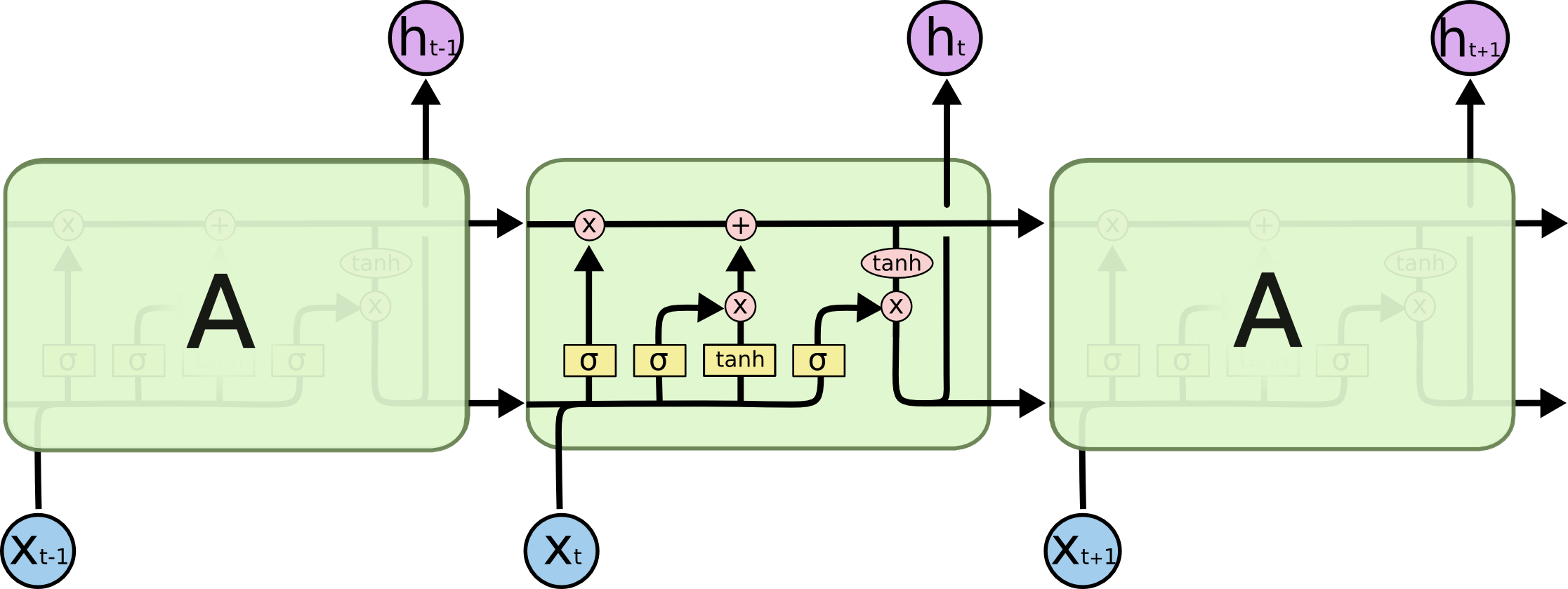}
    \caption{LSTM architecture (courtesy of Christopher Olah)}
    \label{fig:vs3}
\end{figure}

Each LSTM unit (Figure \ref{fig:vs3}) holds a cell state $C$ which flows from timestep $t-1$ to t where some modifications occur in between due to the LSTM gates, these gates are a way to optionally let information pass through. They are composed of a sigmoid layer that outputs a value between 0 and 1 and a multiplication operation in which multiplying by 0 means prevent the information flow and 1 means allow full information flow.
The forget gate receives the hidden state at the previous time step $h_{t-1}$ and the input $x_t$ and decides how much information needs to be forgotten by the following equation:
\begin{equation}
f_t = sigmoid(W_f \thinspace . \thinspace[h_{t-1}, x_t] + b_f)
\end{equation}
The input gate decides what new information need to be stored in the cell state, it follows the following equation:
\begin{equation}
i_t = sigmoid(W_i \thinspace . \thinspace [h_{t-1}, x_t] + b_i)
\end{equation}
We also compute a candidate set of values to be added to the cell state:
\begin{equation}
\tilde{C_t} = tanh(W_c \thinspace . \thinspace [h_{t-1}, x_t] + b_c)
\end{equation}
Following these computations is the update on the cell state which occurs by the equation:
\begin{equation}
C_t = f_t * C_{t-1} + i_t * \tilde{C_t}
\end{equation}
Finally the model decides on what it should output according to its output gate:
\begin{equation}
o_t = sigmoid(W_o \thinspace . \thinspace [h_{t-1}, x_t] + b_o)
\end{equation}
The output to the hidden state is then computed by:
\begin{equation}
o_t = tanh(C_t)
\end{equation}

\begin{figure}[h]
    \centering
    \includegraphics[scale=0.3]{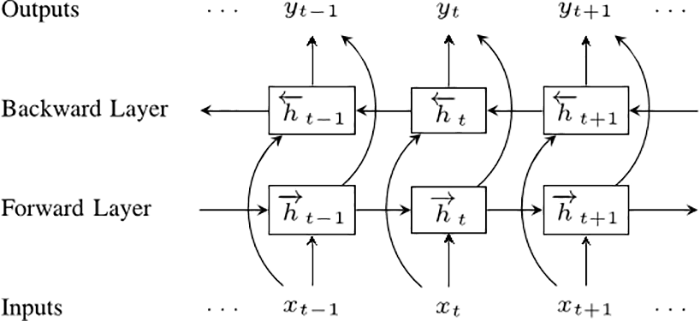}
    \caption{Bi-LSTM architecture \cite{Graves2013HybridSR}}
    \label{fig:bilstm}
\end{figure}
A Bi-directional LSTM layer (Figure \ref{fig:bilstm}) is added to our model instead of the classic Uni-directional LSTM since allowing the model to read text right-to-left and left-to-right and aggregating their outputs would improve it significantly by peeking into the future as well as benefiting from the past \cite{huang2015bidirectional} \cite{lample2016neural}. It will also make the model invariant to language writing direction. for example we train the model on English which is a left-to-right language and evaluate it on Arabic which is a right-to-left language.

We pass the $h_t$ vector for every time step to the next layer which can be another layer of Bi-LSTM for an added layer of feature extraction or a Conditional Random Field classifier.
\subsection{Conditional Random Field Classifier}
Going from the features extracted by the Bi-LSTM to class labels requires a classifier that can take as input the features at each time step and produce as output the class label for each word.

For sequence classification the most common classifiers used in the domain of NER are:
\begin{itemize}
\item LSTM
\item Hidden Markov Model
\item Simple RNN
\item Conditional Random Field
\end{itemize}
The CRF classifier is used for its strength and its current results in state of the art NER models \cite{huang2015bidirectional} \cite{lample2016neural}. A huge strength of CRF is its ability to work with output classes that are dependent, an example of this is that for any I-*(Inside class) there must be a preceding B-*(Beginning class). CRF is capable of exploiting this dependency thus improving the results and avoiding unintuitive class outputs.

Let $h_{1:N}$ be our input to the CRF layer which are the features for each word in a sequence of length N where (1:N) denotes the words from index 1 to N. These features are calculated by $h_t = \text{Bi-LSTM}(x_t)$. let $z_{1:N}$ be our class labels for the input sequence. The linear chain CRF model defines conditional probability of the class labels given the input by the equation:
\begin{equation}
p(z_{1:N} | h_{1:N}) = \frac{1}{Z} \thinspace exp \thinspace (\sum_{n=1}^{N}\sum_{i=1}^{F}\lambda_if_i(z_{n-1},z_n, h_{1:N}, n))
\end{equation}
In the equation the term Z is the normalization factor or the partition function which is used to make $p(z_{1:N} | h_{1:N})$ a valid probability, the term F is the number of feature functions we're going to use.
\begin{equation}
Z = \sum_{z_1:N} exp \thinspace (\sum_{n=1}^N \sum_{i=1}^F\lambda_if_i(z_{n-1},z_n, h_{1:N}, n))
\end{equation}
In the probability equation we had feature functions $f_i$, each of which is a function that takes as input the current class label, the label of the previous class, the input and the current position and outputs a boolean value 0 or 1.
$\lambda_i$ is the feature weight for feature $i$, if this weight is large and positive then we emphasis that the probability of the current label is high if this feature is true. An example of a feature function: $f_1(z_{n-1}, z_n, h_{1:N}, n) = 
\begin{cases}
       \text{1,} \quad\text{if } z_{n-1} = \text{B-PERS} \text{ and } h_t = \text{Bi-LSTM}(\text{John})\\
       \text{0,} \quad\text{Otherwise} \\ 
\end{cases}
$
A positive weight for this function would indicate that the model prefers the tag B-PERS for the current word, if the function was assigned a negative weight it would indicate that the model does not prefer the tag B-PERS for the current word.

In our case the features will be the output of the Bi-LSTM. In this case the Bi-LSTM layer(s) can be considered as the feature function we are using.

The appropriate objective for parameter learning of CRF is to maximize the conditional likelihood of the training data where m is the number of training examples:
\begin{equation} \label{eq:crf-obj}
\sum_{j=1}^{m}{log \ p(z_{1:N} | h_{1:N})}
\end{equation}
This learning procedure is conducted by computing the gradients for the objective function (Equation \ref{eq:crf-obj}) and use the gradient in a gradient based optimization algorithm like Stochastic Gradient Descent.

Now that the model is trained we can use it to calculate the probability $p(z_{1:N} | h_{1:N})$ for any tag for each token in an input sequence. An approach to take the final sequence is by greedily taking the tag that has the highest probability. However this approach would have to check an exponential number of tags because it has to check $K^N$ possible tags with K being the number of classes in our model. Another approach would be to use the Viterbi algorithm \cite{forney1973viterbi} which is a dynamic programming algorithm that works in polynomial time to find the optimal sequence of tags.

\section{Experimental Results}
\label{sec5}
First we define the metrics we used in our evaluation. We used Precision and Recall and their harmonic mean (F1-score).

Precision is defined as the number of true positives divided by the number of true positives and false positives. It is considered as a measure of the classifier's exactness.
\begin{equation}
Precision = \frac{True Positives}{True Positives + False Positives}
\end{equation}
Recall is defined as the number of true positives divided by the number of true positives and false negatives. It can be considered the sensitivity or the true positive rate.
\begin{equation}
Recall = \frac{True Positives}{True Positives + False Negatives}
\end{equation}
In our evaluation we use the F1-score which is a balance between them.
\begin{equation}
F1 score = 2 * \frac{Precision * Recall}{Precision + Recall}
\end{equation}
We trained our model on the WikiNER English dataset \cite{nothman2012:artint:wikiner} by using the word embedding model \cite{bojanowski2016enriching} which is open source. We evaluated the model on English and noted its results then we tested it on the ANER dataset \cite{benajiba2007anersys} with only applying the transformation and no fine tuning of the model.

The model was written using Keras \cite{chollet2015keras}, it was trained on an Ubuntu 16.04 system running on an Intel i7-7500U processor and accelerated by an Nvidia Geforce 940MX GPU. All the code for the experiment is available as open source on Github: \href{https://github.com/Omarito2412/language-independent-ner}{Code on Github}

The results of the model imply that the model succeeded at detecting many named entities in the Arabic text without being exposed to Arabic before. This approach is valid for any other language that we have a transformation matrix for. Recently 78 matrices for transformation \cite{smith2017offline} have been released as open source. These matrices can be used to transform the word embedding space from any of the 78 languages to English, which can be used to detect named entities in 78 different languages with only being trained on English.

\begin{table}[!h]
\centering
\begin{tabular}{|c|c|c|c|}
\hline
\multicolumn{4}{|c|}{Classification results per class for English} \\
\hline
& Precision & Recall & F1-Score \\
\hline
PER & 0.88 & 0.93 & 0.91\\
\hline
MISC & 0.80 & 0.73 & 0.76\\
\hline
LOC & 0.86 & 0.85 & 0.86\\
\hline
ORG & 0.82 & 0.72 & 0.76\\
\hline
Average scores & 0.84 & 0.82 & 0.83\\
\hline
\end{tabular}
\caption{Classification Report for Precision, Recall and F1-Scores on WikiNER}
\label{table:1}
\end{table}

We can see from Table \ref{table:1} that the model successfully learns to detect named entities in the English text. The model's accuracy depends on many factors among which the training data and the complexity of the model. The model seems to perform best on the persons class and seems to perform worst on the miscellaneous class.

\begin{table}[!h]
\centering
\begin{tabular}{|c|c|c|c|}
\hline
\multicolumn{4}{|c|}{Classification results per class for Arabic} \\
\hline
& Precision & Recall & F1-Score \\
\hline
PER & 0.05 & 0.01 & 0.01\\
\hline
MISC & 0.01 & 0.57 & 0.02\\
\hline
LOC & 0.07 & 0.00 & 0.00\\
\hline
ORG & 0.02 & 0.05 & 0.03\\
\hline
Average scores & 0.05 & 0.07 & 0.02\\
\hline
\end{tabular}
\caption{Classification Report for Precision, Recall and F1-Scores on ANER without Alignment}
\label{table:2}
\end{table}

In Table \ref{table:2} the model is tested on ANER which is the Arabic dataset without aligning the word vectors with the English word vectors. Recall that the model was never trained to understand Arabic and so testing the model on Arabic would produce a score of 0 (or almost 0) which is the case as seen in the table. The model fails to capture any entities -Except for a very tiny number of entities that could be random-.

\begin{table} [!h]
\centering
\begin{tabular}{|c|c|c|c|}
\hline
\multicolumn{4}{|c|}{Classification results per class for Arabic} \\
\hline
& Precision & Recall & F1-Score \\
\hline
PER & 0.68 & 0.52 & 0.59\\
\hline
MISC & 0.05 & 0.06 & 0.06\\
\hline
LOC & 0.80 & 0.44 & 0.57\\
\hline
ORG & 0.32 & 0.08 & 0.12\\
\hline
Average scores & 0.58 & 0.36 & 0.43\\
\hline
\end{tabular}
\caption{\label{table:classification-report} Classification Report for Precision, Recall and F1-Scores on ANER with Alignment}
\label{table:3}
\end{table}

The results in Table \ref{table:3} shows that after alignment the model succeeds at detecting named entities in Arabic which it was never trained on. It detects the class persons with precision of 68$\%$ compared to 88$\%$ in English and it's also its top scoring class.

\begin{figure}[!htb]
   \begin{minipage}{0.4\textwidth}
    \centering
    \includegraphics[scale=0.4]{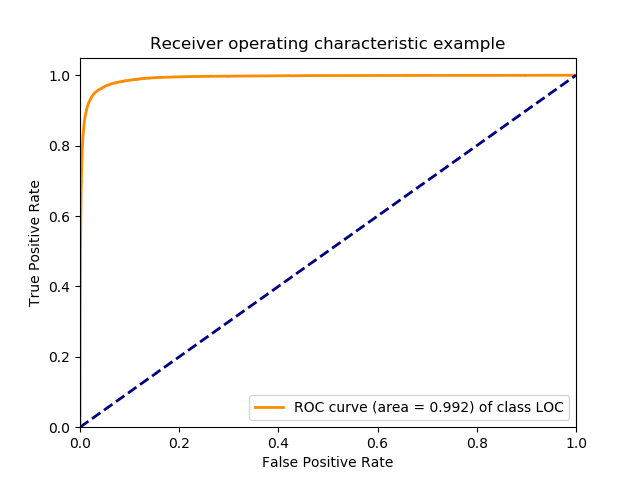}
    \caption{ROC curve for LOC class in English}
    \label{fig:LOC_roc}
    \end{minipage}\hfill \hspace{1cm}
   \begin {minipage}{0.4\textwidth}
    \centering
    \includegraphics[scale=0.4]{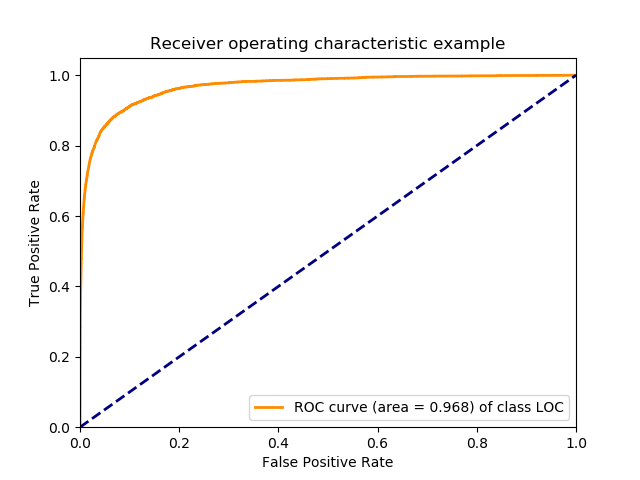}
    \caption{ROC curve for LOC class in Arabic}
    \label{fig:LOC_AR_roc}
   \end{minipage}
\end{figure}

\begin{figure}[!htb]
   \begin{minipage}{0.4\textwidth}
    \centering
    \includegraphics[scale=0.4]{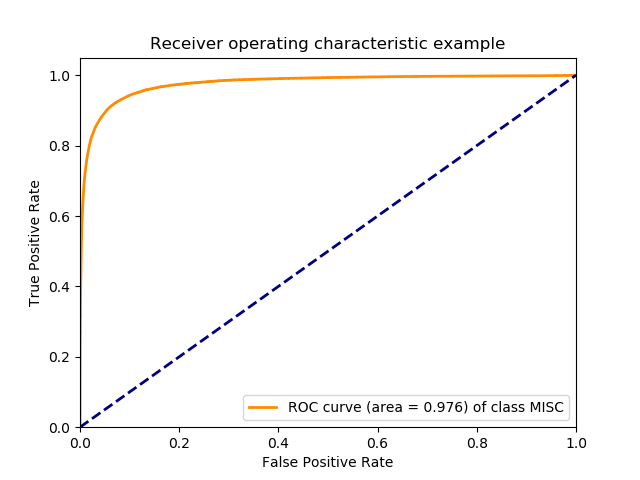}
    \caption{ROC curve for MISC class in English}
    \label{fig:MISC_roc}
    \end{minipage}\hfill \hspace{1cm}
   \begin {minipage}{0.4\textwidth}
    \centering
    \includegraphics[scale=0.4]{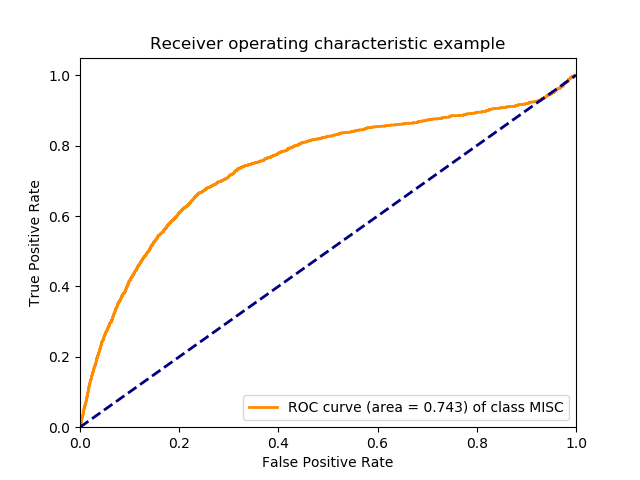}
    \caption{ROC curve for MISC class in Arabic}
    \label{fig:MISC_AR_roc}
   \end{minipage}
\end{figure}

\begin{figure}[!htb]
   \begin{minipage}{0.4\textwidth}
    \centering
    \includegraphics[scale=0.4]{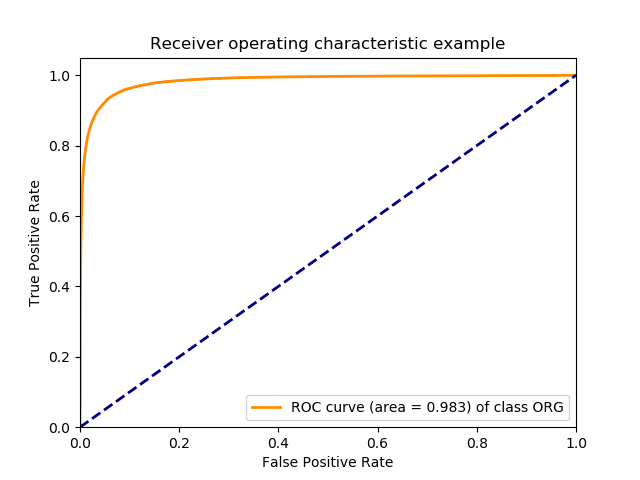}
    \caption{ROC curve for ORG class in English}
    \label{fig:ORG_roc}
    \end{minipage}\hfill \hspace{1cm}
   \begin {minipage}{0.4\textwidth}
    \centering
    \includegraphics[scale=0.4]{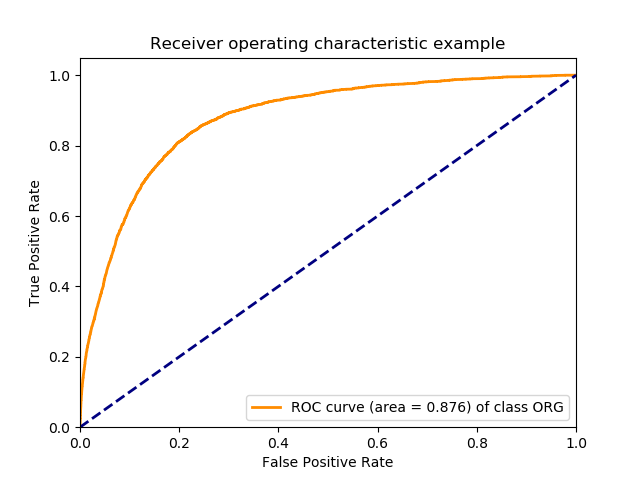}
    \caption{ROC curve for ORG class in Arabic}
    \label{fig:ORG_AR_roc}
   \end{minipage}
\end{figure}

\begin{figure}[!htb]
   \begin{minipage}{0.4\textwidth}
    \centering
    \includegraphics[scale=0.4]{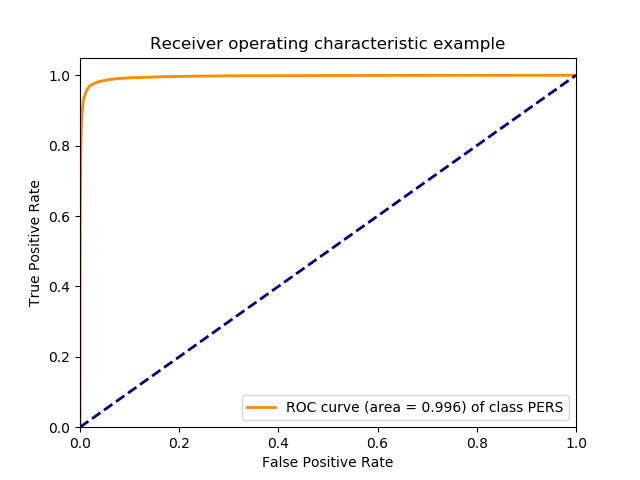}
    \caption{ROC curve for PERS class in English}
    \label{fig:PERS_roc}
    \end{minipage}\hfill \hspace{1cm}
   \begin {minipage}{0.4\textwidth}
    \centering
    \includegraphics[scale=0.4]{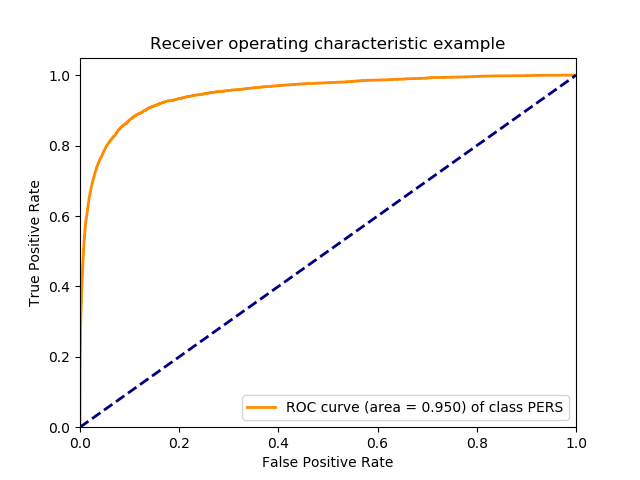}
    \caption{ROC curve for PERS class in Arabic}
    \label{fig:PERS_AR_roc}
   \end{minipage}
\end{figure}

ROC figures and their corresponding AUC in Figures \ref{fig:LOC_roc}, \ref{fig:LOC_AR_roc}, \ref{fig:MISC_roc}, \ref{fig:MISC_AR_roc}, \ref{fig:ORG_roc}, \ref{fig:ORG_AR_roc}, \ref{fig:PERS_roc}, \ref{fig:PERS_AR_roc} show that the proposed approach performs very well compared to random guessing and that the system is capable of classifying the aforementioned classes in both English and Arabic without the need for retraining the model on Arabic.

\begin{figure}[!htb]
   \begin {minipage}{0.38\textwidth}
    \centering
    \includegraphics[scale=0.38]{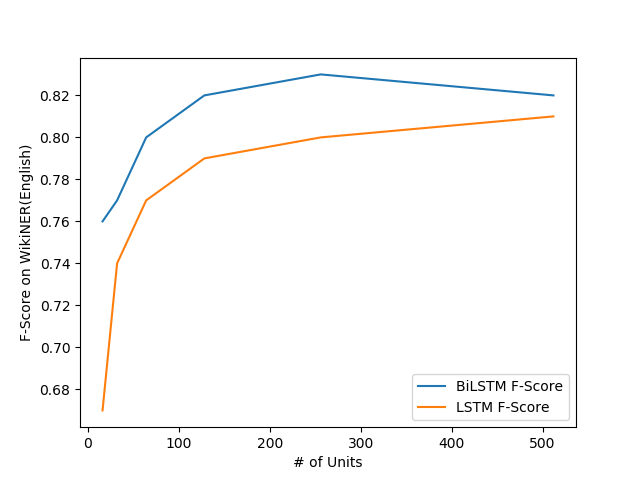}
    \caption{LSTM vs BiLSTM performance \\ on English data}
    \label{fig:bilstm-vs-lstm-english}
   \end{minipage}
   \begin{minipage}{0.38\textwidth}
    \centering
    \includegraphics[scale=0.38]{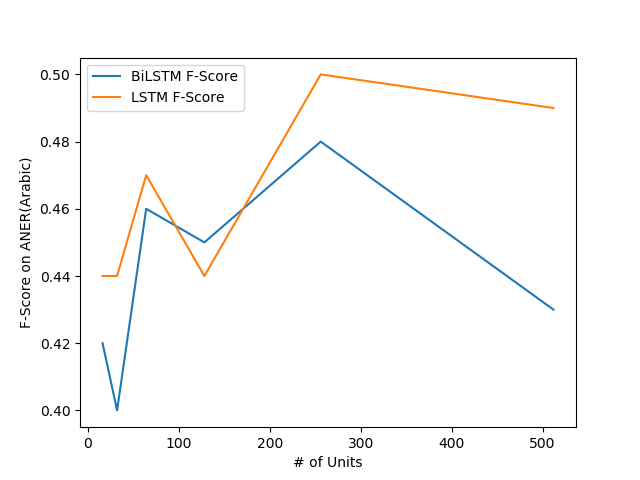}
    \caption{LSTM vs BiLSTM performance on aligned Arabic data}
    \label{fig:bilstm-vs-lstm-arabic}
    \end{minipage}\hfill \hspace{1cm}
\end{figure}

Hyper parameter selection for the model was performed empirically by iterating over a finite set of parameters and choosing values that maximize our objective function (Grid search).

The main set of parameters to choose from were whether to use Bidirectional LSTM or Unidirectional LSTM and the number of units at the LSTM layer. These parameters were tested on both the English and Arabic models.

Figure \ref{fig:bilstm-vs-lstm-english} shows that BiLSTM performance is superior to LSTM performance on English data while it fluctuates on Arabic data according to Figure \ref{fig:bilstm-vs-lstm-arabic}. It's also noted that the F-score values starts degrading just below 300 units. This makes the optimal choice for units to be around 256 units.
% \begin{figure}[h]
%     \centering
%     \includegraphics[scale=0.5]{arabic_bilstm_result}
%     \caption{BiLSTM performance on Arabic data}
%     \label{fig:bilstm-arabic}
% \end{figure}
% \begin{figure}[h]
%     \centering
%     \includegraphics[scale=0.5]{arabic_lstm_result}
%     \caption{LSTM performance on Arabic data}
%     \label{fig:lstm-arabic}
% \end{figure}

% \begin{figure}[h]
%     \centering
%     \includegraphics[scale=0.5]{english_bilstm_result}
%     \caption{BiLSTM performance on English data}
%     \label{fig:bilstm-english}
% \end{figure}
% \begin{figure}[h]
%     \centering
%     \includegraphics[scale=0.5]{english_lstm_result}
%     \caption{LSTM performance on English data}
%     \label{fig:bilstm-english}
% \end{figure}

\section{Conclusion}
\label{sec6}
We exploit orthogonal transformation of word embeddings to create a language independent NER model that was trained on English and evaluated on Arabic without being explicitly exposed to Arabic before or fine tuned on an Arabic dataset. This paves the road to language independent NLP models that are capable of solving typical NLP tasks without being explicitly trained on a specific language.
\clearpage
\bibliography{refs}
\bibliographystyle{unsrt}
\end{document}